\documentclass[letterpaper]{article} % DO NOT CHANGE THIS
\usepackage{aaai2026}  % DO NOT CHANGE THIS
\usepackage{times}  % DO NOT CHANGE THIS
\usepackage{helvet}  % DO NOT CHANGE THIS
\usepackage{courier}  % DO NOT CHANGE THIS
\usepackage[hyphens]{url}  % DO NOT CHANGE THIS
\usepackage{graphicx} % DO NOT CHANGE THIS
\usepackage{natbib}  % DO NOT CHANGE THIS AND DO NOT ADD ANY OPTIONS TO IT
\usepackage{caption} % DO NOT CHANGE THIS AND DO NOT ADD ANY OPTIONS TO IT
\usepackage{amsfonts}       % blackboard math symbols
\usepackage{amsthm,amsmath,amssymb}
\usepackage{mathrsfs}
\usepackage{booktabs}
\usepackage{framed}

\usepackage{algorithm}
\usepackage{algorithmic}

\usepackage{newfloat}
\usepackage{listings}
\DeclareCaptionStyle{ruled}{labelfont=normalfont,labelsep=colon,strut=off} % DO NOT CHANGE THIS
\floatstyle{ruled}
\newfloat{listing}{tb}{lst}{}
\floatname{listing}{Listing}
\nocopyright

\title{A3D: \underline{A}daptive \underline{A}ffordance \underline{A}ssembly with \underline{D}ual-Arm Manipulation}
\author{
    Jiaqi Liang\textsuperscript{\rm 1}\equalcontrib, 
    Yue Chen\textsuperscript{\rm 1}\equalcontrib,
    Qize Yu\textsuperscript{\rm 1}\equalcontrib,
    Yan Shen\textsuperscript{\rm 1},
    Haipeng Zhang\textsuperscript{\rm 1},
    Hao Dong\textsuperscript{\rm 1},
    Ruihai Wu\textsuperscript{\rm 1}\thanks{Corresponding Authors.}
}
\affiliations{
    \textsuperscript{\rm 1}Peking University

    \fontsize{10pt}{0}{jiaqiliang@stu.pku.edu.cn, 
    yuechen@stu.pku.edu.cn, 
    wuruihai@pku.edu.cn}
}

\usepackage{bibentry}
\begin{document}

\maketitle

\begin{abstract}
Furniture assembly is a crucial yet challenging task for robots, requiring precise dual-arm coordination where one arm manipulates parts while the other provides collaborative support and stabilization. 
To accomplish this task more effectively, robots need to actively adapt support strategies throughout the long-horizon assembly process, while also generalizing across diverse part geometries. 
We propose A3D, a framework which learns adaptive affordances to identify optimal support and stabilization locations on furniture parts. The method employs dense point-level geometric representations to model part interaction patterns, enabling generalization across varied geometries. To handle evolving assembly states, we introduce an adaptive module that uses interaction feedback to dynamically adjust support strategies during assembly based on previous interactions.
We establish a simulation environment featuring 50 diverse parts across 8 furniture types, designed for dual-arm collaboration evaluation. Experiments demonstrate that our framework generalizes effectively to diverse part geometries and furniture categories in both simulation and real-world settings.
\end{abstract}

\begin{links}
    \link{Code}{https://github.com/a3d-11011/Adaptive-Affordance-Assembly-with-Dual-Arm-Manipulation}
\end{links}

\begin{figure}[!ht]
\centering
\includegraphics[width=\linewidth]{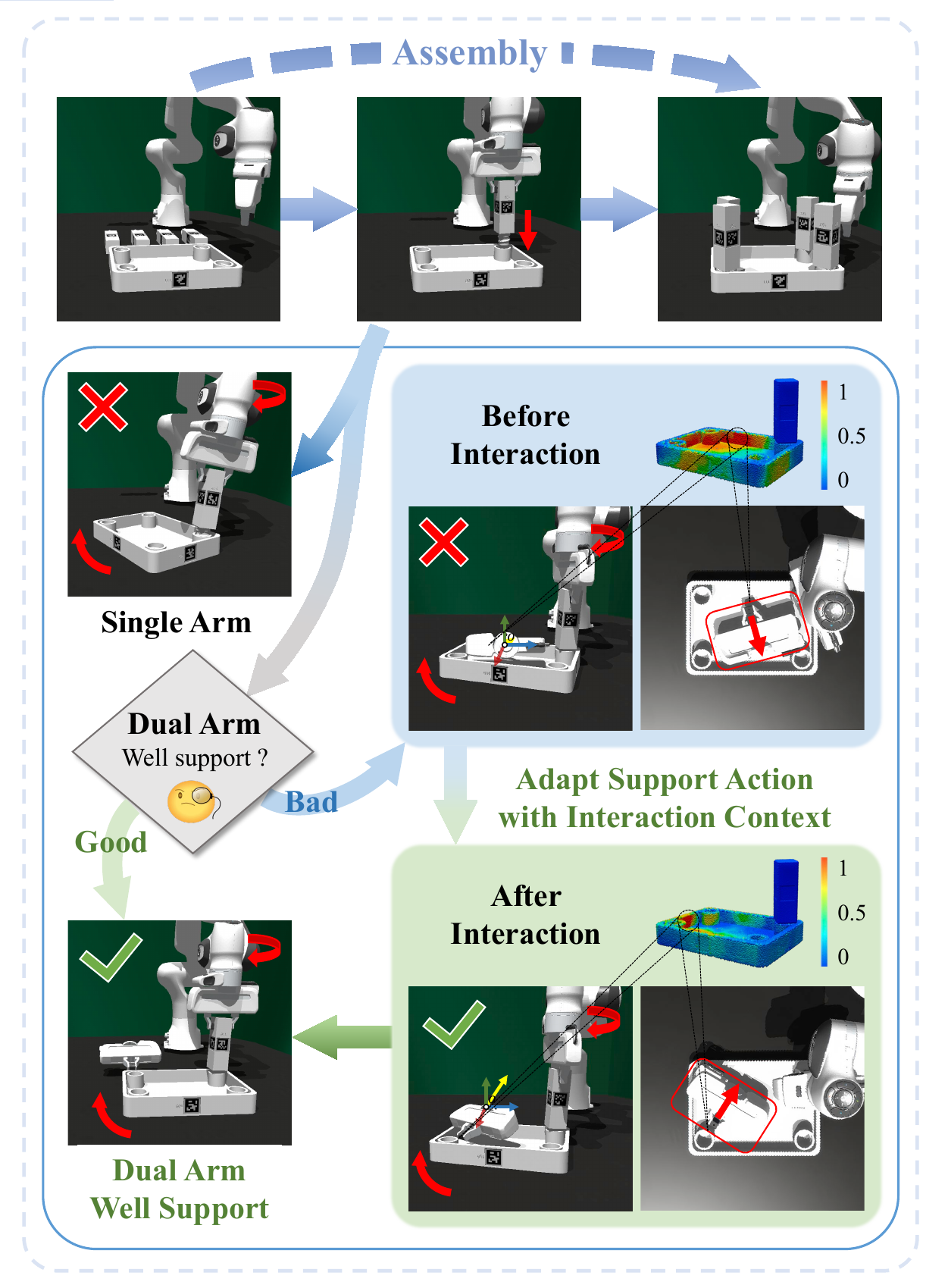} % Reduce the figure size so that it is slightly narrower than the column.
\caption{
Procedure of assembling a furniture (Row 1).
\textbf{Single Arm} may not stably assemble parts and a second robot is then introduced.
\textbf{Before Interaction}, part kinematics and dynamics, indicated by affordance, are ambiguous, and the interaction may fail.
\textbf{After Interaction}, the adapted affordance proposes actions for stable support during assembly.
}
\label{teaser}
\end{figure}

\section{Introduction}
Robotic furniture assembly~\citep{funkhouser2011learning,jones2021automate,lee2021ikea,tian2022assemble,tian2025fabric}, the task of combining functional components such as chair base, legs, and arms into a fully constructed shape, with a focus on both the overall structure and functions of each part, is a critical capability for home-assistive robots.

Recent studies have addressed various aspects of robotic assembly, including motion planning~\citep{suarez2018can,sundaram2001disassembly,le2009path,zhang2020c}, assembly pose estimation~\citep{yu2021roboassembly,huang2020generative,tie2025manual2skill,jones2021automate, shen2025biassemble}, and RL-based combinatorial sequence search~\citep{xu2023efficient,zhang2024bridging,funk2022learn2assemble,ghasemipour2022blocks}. However, current robotic systems remain limited in their ability to assemble objects across diverse categories. Prior research has primarily focused on specific object types using a single robot arm~\citep{heo2023furniturebench,lee2021ikea}. In contrast, generalizable furniture assembly requires vision understanding and bi-manual operation that frequently changes which part to hold to counter-balance the insertion force from the other hand. This presents new challenges to vision perception and precise manipulation. First, assembling unseen furniture demands understanding functional affordances across various part geometries, requiring robots to identify viable support locations. Second, long-horizon assembly induces sequential state transitions where support strategies must dynamically refine based on part relations. Third, fine-grained assembly requires robust control skills to achieve precisly during contact-rich interactions.

To bridge these gaps, we introduce \textbf{A3D}: a framework that learns \textbf{A}daptive \textbf{A}ffordance \textbf{A}ssembly for collaborative \textbf{D}ual-arm manipulation. To enable geometric awareness, A3D leverages affordance as a representation of per-point actionability on objects for furniture assembly tasks. These per-point features are extracted hierarchically from local to global, effectively capturing detailed local geometry information for support and stabilization, as well as the contextual part relations that indicate whether the action would disturb other parts. This hierarchical structure enables A3D to localize stable support regions through fine-grained geometric cues while modeling higher-level part contexts to anticipate potential disturbances during manipulation.

However, static affordance derived solely from passive observations fails to account for critical kinematic (e.g.,  joint locations and limits) and dynamic uncertainties (e.g., contact direction and force), which might misdirect manipulation (Fig.~\ref{teaser}). So we actively incorporate interaction feedback into affordance predictions, enabling dynamic adjustment of support strategies throughout assembly.

% Furthermore, the absence of suitable simulation environments has significantly obstructed progress in dual-arm furniture assembly research. 
Although existing simulation environments have facilitated progress in robotic manipulation, they remain limited in supporting the study of dual-arm furniture assembly. 
Previous works predominantly focus on single-arm manipulation or utilize limited furniture assets ~\citep{niekum2013incremental, suarez2018can, kimble2020benchmarking, heo2023furniturebench, zhang2024bridging, jones2021automate, li2020learning}, failing to capture the unique physical and coordination challenges inherent in dual-arm assembly scenarios. To bridge this critical gap, we introduce a new evaluation environment extending FurnitureBench~\citep{heo2023furniturebench}, featuring 4 assembly task categories with 50 geometrically diverse parts across 8 furniture types. Both qualitative and quantitative results from simulations and real-world experiments demonstrate the effectiveness of our framework.
We also note that while our adaptation allows iterative refinement (max 3 rounds), most test cases succeeded after a single interaction (effective $k=1$), demonstrating robustness and efficiency.

In conclusion, our contributions mainly include:

\begin{itemize}
    \item We propose affordance learning framework for generalizable support and stabilization prediction in furniture assembly, enabling generalization across diverse parts.
    
    \item We further develop an adaptive module that uses interaction feedback to dynamically adjust support strategies during assembly based on previous interactions.

    \item We build a simulation environment for dual-arm collaborative assembly featuring 50+ geometrically diverse parts across 8 furniture types and 4 task categories.
    
    \item Extensive experiments in both simulation and real world demonstrate the effectiveness of our framework.
\end{itemize}

\section{Related Work}
% These
\subsection{Furniture Assembly}
Furniture assembly is a prominent application in shape assembly, where individual components, each serving a distinct functional role (\emph{e.g.}, chair arm, table leg), must be assembled following both geometric constraints and common-sense spatial and functional relations (e.g., a chair leg must be attached to the seat base with proper orientation and stability). The complexity arises from the need to reason about part functionality, structural dependencies, and physical constraints simultaneously. Previous research has mostly focused on assembly pose estimation~\citep{li2020learning,yu2021roboassembly,li2024category,huang2020generative}. For instance, ~\citet{li2020learning} learns to assemble 3D shapes from 2D images, while \citet{huang2020generative} proposes image-free generative models for pose generation. 
However, these methods might neglect the challenges in dynamic robotic execution, particularly the need for precise dual-arm coordination, where one arm manipulates one part while the other actively provides collaborative support and stabilization throughout assembly. Addressing this challenge,
% in real-world robotic furniture assembly, 
especially adaptive support strategies and generalization across diverse geometries, is a core objective of our work.

\subsection{Visual Affordance for Robotic Manipulation}
Visual affordance~\cite{gibson1977theory} suggests possible ways for
agents to interact with objects for various manipulation tasks. This approach has been widely used in grasping~\cite{corona2020ganhand, kokic2020learning, zeng2018robotic}, articulated manipulation~\cite{yuan2024general, tie2025etseed}, and scene interaction~\cite{interaction-exploration, ego-topo}. Point-level affordance, in particular, assigns an actionability score to each point, and thus enables fine-grained geometry understanding and improved cross-shape generalization in diverse tasks, such as articulated~\cite{mo2021where2act, wang2021adaafford, chen2024eqvafford}, and deformable~\cite{Wu_2024_CVPR, wu2023learning, wu2025garmentpile, wang2025dexgarmentlab} manipulation. For furniture assembly scenarios, where parts vary significantly in geometry and require precise dual-arm collaboration, we empower point-level affordance with the awareness of part geometry, and further leverage active interactions to efficiently query uncertain kinematic or dynamic factors for learning more accurate instance-adaptive visual affordance.

\begin{figure*}[t]
\centering
\includegraphics[width=\textwidth]{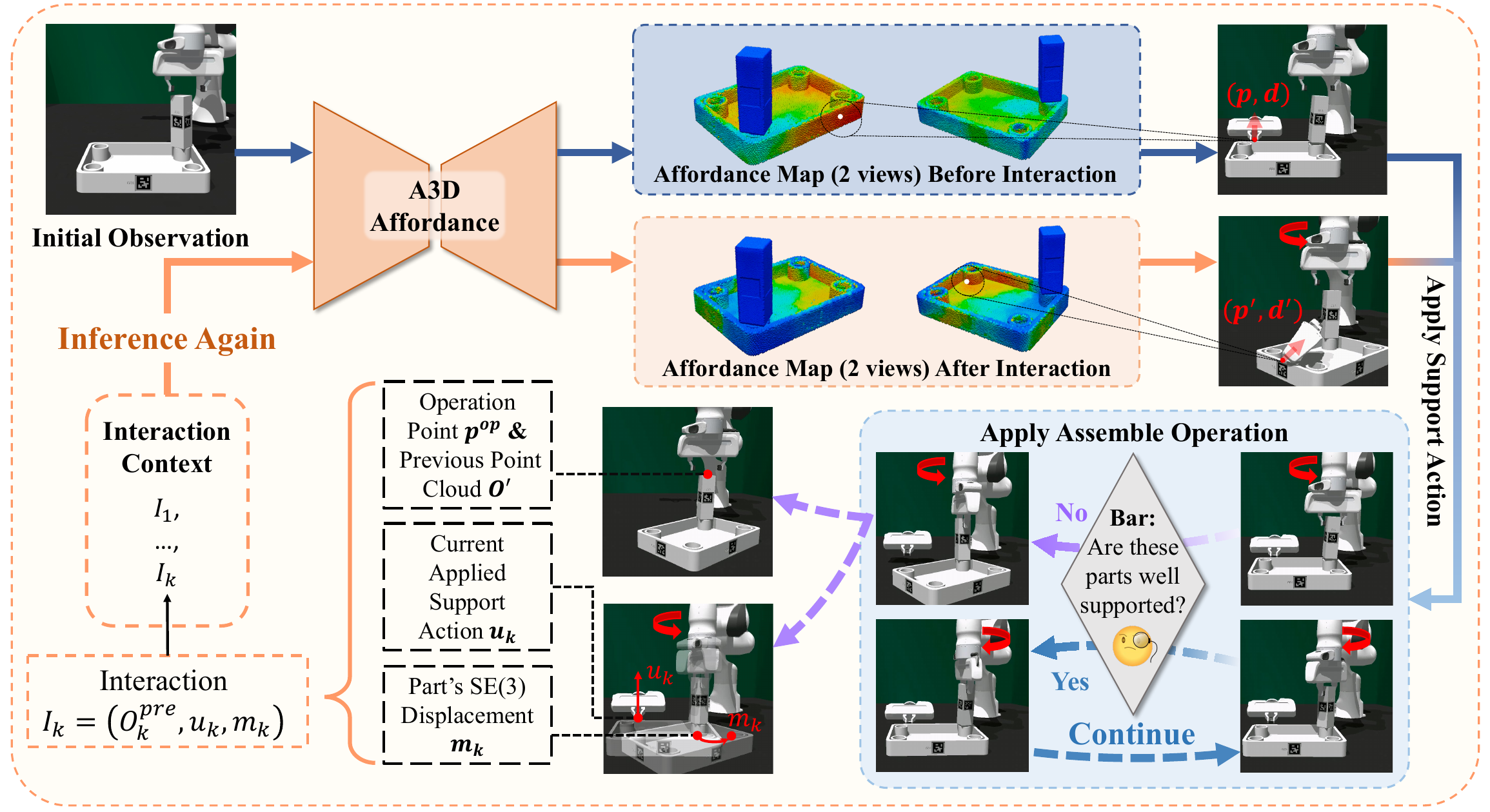} % Reduce the figure size so that it is slightly narrower than the column.
\caption{\textbf{Framework Overview.} At each operation stage, the policy takes the point cloud and the selected action point as inputs to predict the support action. The robot moves the gripper to the recommended pose to support the assembly. If part displacement occurs—indicating insufficient support—the system logs the pre-support point cloud, executed action, and displacement as interaction context, then re-predicts the support action using the updated point cloud and accumulated context.}
\label{fig:framework}
\end{figure*}

\begin{figure*}[t]
\centering
\includegraphics[width=\textwidth]{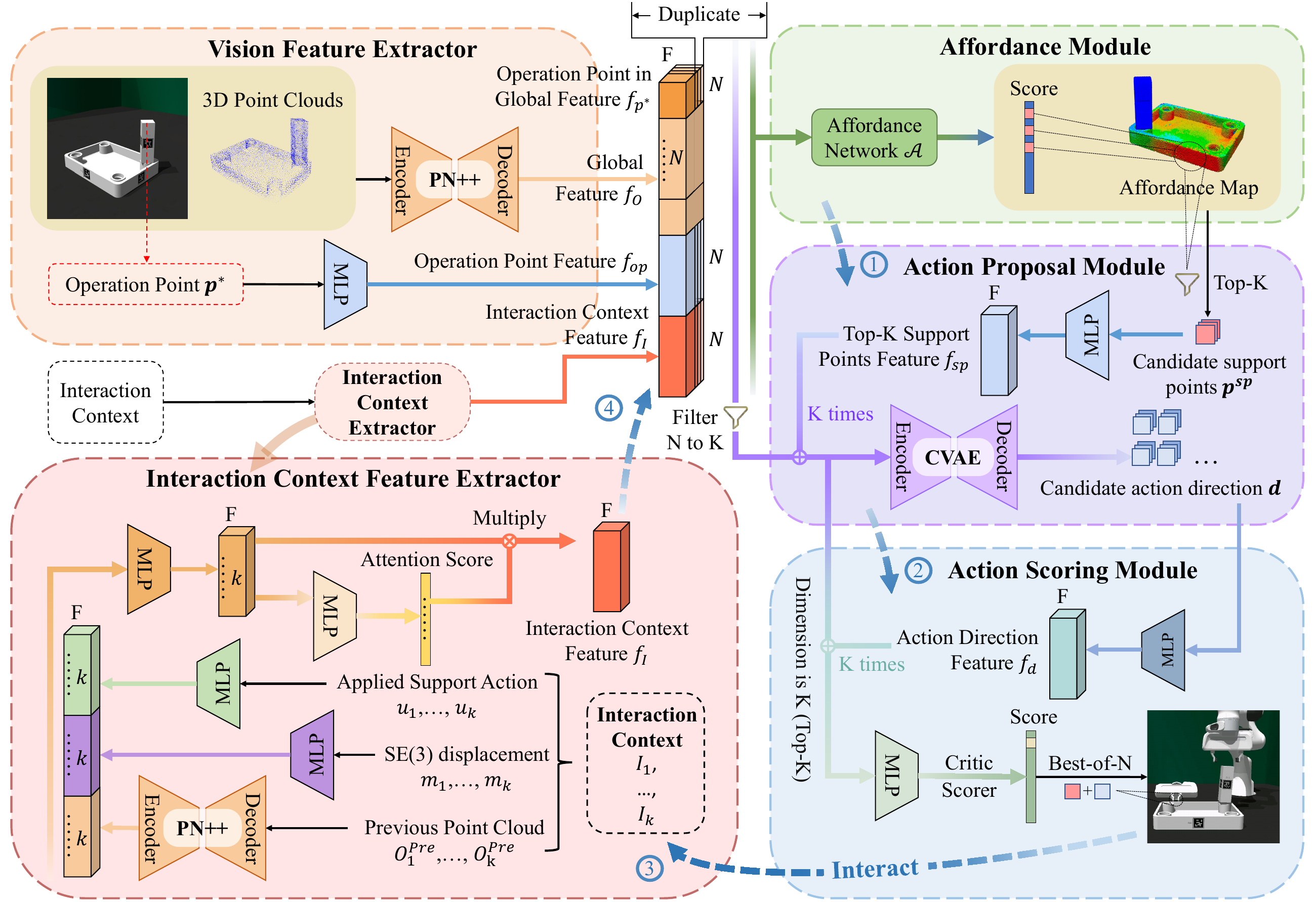} % Reduce the figure size so that it is slightly narrower than the column.
\caption{\textbf{Point-Level Adaptation Support Affordance Framework.} 
The model completes support decisions by extracting visual features, computing Top-K point-level affordances and generating candidate directions, scoring and selecting point–direction pairs, and extracting interaction context features.
}
\label{pipline}
\end{figure*}

\section{Method}
Our goal is to enable effective dual-arm coordination for furniture assembly, where a tool arm executes assembly operations while a support arm provides adaptive stabilization to prevent part displacement and ensure task success. As shown in Fig.~\ref{fig:framework}, our framework integrates two key components: (1) \textbf{Support Affordance Module} predicts initial affordance heatmaps and corresponding action directions from visual observations and operation points; (2) \textbf{Interaction Context Adaptation Module} leverages physical feedback from interaction history to adjust affordance predictions.

\subsection{Problem Formulation}\label{sec:formulation}
We formulate this as learning a closed-loop adaptive policy $\pi(u_t|S_t, I_t)$. At each timestep $t$, the policy predicts the stabilization action $u_t$ for the support arm, conditioned on the observed state $S_t$, and the interaction context $I_t$ which records the history of previous assembly trials. 

\textbf{State}: $S_t = (O_t, p^{op}_t)$, where $O_t \in \mathbb{R}^{N\times6}$ represents a 3D partial point cloud of the furniture parts with surface normals, and $p^{op}_t$ denotes the operation point where the tool gripper contacts the target part.

\textbf{Action}: $u_t=(p^{sp}_t,\mathbf{d_t})$, where $p^{sp}_t \in O_t$ is the support point and $\mathbf{d_t} \in SO(3)$ is the support gripper orientation.

\textbf{Interaction Context}: 
$I_t = \{ (O_i, u_i, m_i) \}_{i = t-k}^{t-1}$ stores information from previous $k$ interaction steps, where $m_i$ denotes the base part displacement after step $is$.

\textbf{Task Success}:
An episode succeeds if the primary operation reaches its geometric goal while maintaining base part displacement $m_i < \epsilon$ during execution.

\subsection{Support Affordance Module}

The Support Affordance Module employs an affordance–proposal–scoring architecture: the Affordance submodule predicts affordance maps and selects top-K candidate points; the Proposal submodule generates multiple candidate directions for each point; the Scoring submodule scores all point–direction pairs and selects the optimal support action (Steps 1–3, Fig. ~\ref{pipline}). 

\subsubsection{Visual Feature Extractor.}
PointNet++~\citep{qi2017pointnet++} generates point-wise features $f_{p_i}\in \mathbb{R}^{128}$ from the point cloud $O$. Operation and support points are encoded via shared MLPs into $f_{op}, f_{sp} \in \mathbb{R}^{32}$, while gripper direction $\mathbf{d}$ and displacement $m$ are encoded into $f_{\mathbf{d}}, f_m \in \mathbb{R}^{32}$.

\subsubsection{Affordance Module.}
Module $\mathcal{A}$ predict an affordance score $a_p \in [0,1]$ for each point $p$. It concatenates the operation-point feature $f_{p^{op}}$, the point feature $f_{p_i}$ the operation-point encoding $f_{op}$, and interaction context $f_I$, feeds this into the MLP, and outputs $a_{p_i}$. The top-K points by score are then selected as support candidates.

\subsubsection{Action Proposal Module.}
Action Proposal Module $\mathscr{P}$ implements a Conditional Variational Autoencoder (cVAE). The encoder processes the operation point feature $f_{p^{op}}$, candidate support point feature $f_{p^{sp}}$ in point cloud features, operation point embedding $f_{op}$, support point embedding $f_{sp}$, and interaction context feature $f_I$, to output latent vector $z\in \mathbb{R}^{128}$. 
The decoder then generates direction vector $\mathbf{d}$ from $z$.

\subsubsection{Action Scoring Module.}
Action Scoring Module $\mathscr{S}$ predicts success scores $c \in [0,1]$ for each action. An MLP takes concatenated features $f_{p^{op}}$, $f_{p^{sp}}$, $f_{op}$, $f_{sp}$, and $f_{\mathbf{d}}$ and outputs the success likelihood. A higher $c$ suggests a greater chance for the support hand to collaborate effectively and complete the task. 

\subsection{Interaction Context Adaptation Module}
% When predictions based solely on visual priors are insufficient to support an action (as shown in step 4 of fig. ~\ref{pipline}), the system records the support action and its feedback. The Interaction Context Extractor then derives context features from these records and concatenates them with the current visual features, feeding the combined representation back into the Affordance, Actor, and Critic submodules to produce support results that adhere to physical dynamics.
If visual priors are insufficient (Fig. ~\ref{pipline}, step 4), the system records the support action and its feedback. The Context Extractor derives features from these records, concatenates them with current visual features, and feeds them back to the Affordance, Proposal, and Scoring submodules to refine predictions that adhere to physical dynamics.

\subsubsection{Interaction Context Extractor Module.}

For interaction context $I_t = { (O_i, u_i, m_i) }_{i = t-k}^{t-1}$, we extract features for each historical step using the same encoders as above. Features are combined via 
\begin{equation}
f_{I_i} = MLP(concat(f_{O_i}, f_{u_i}, f_{m_i})), i \in [t-k, t-1].
\end{equation}
To aggregate information from all previous interactions, we adopt a lightweight attention mechanism. 
Each previous interaction feature $f_{I_i}$ is passed through an MLP to compute an attention weight $w_i$, and the final interaction context feature is obtained as a weighted average:
\begin{equation}
     f_{I}=\frac{\sum_{i=t-k}^{t-1} f_{I_{i}} \times w_{i}}{\sum_{i=t-k}^{t-1} w_{i}}.
\end{equation}

\begin{figure*}[!ht]
\centering
\includegraphics[width=\textwidth]{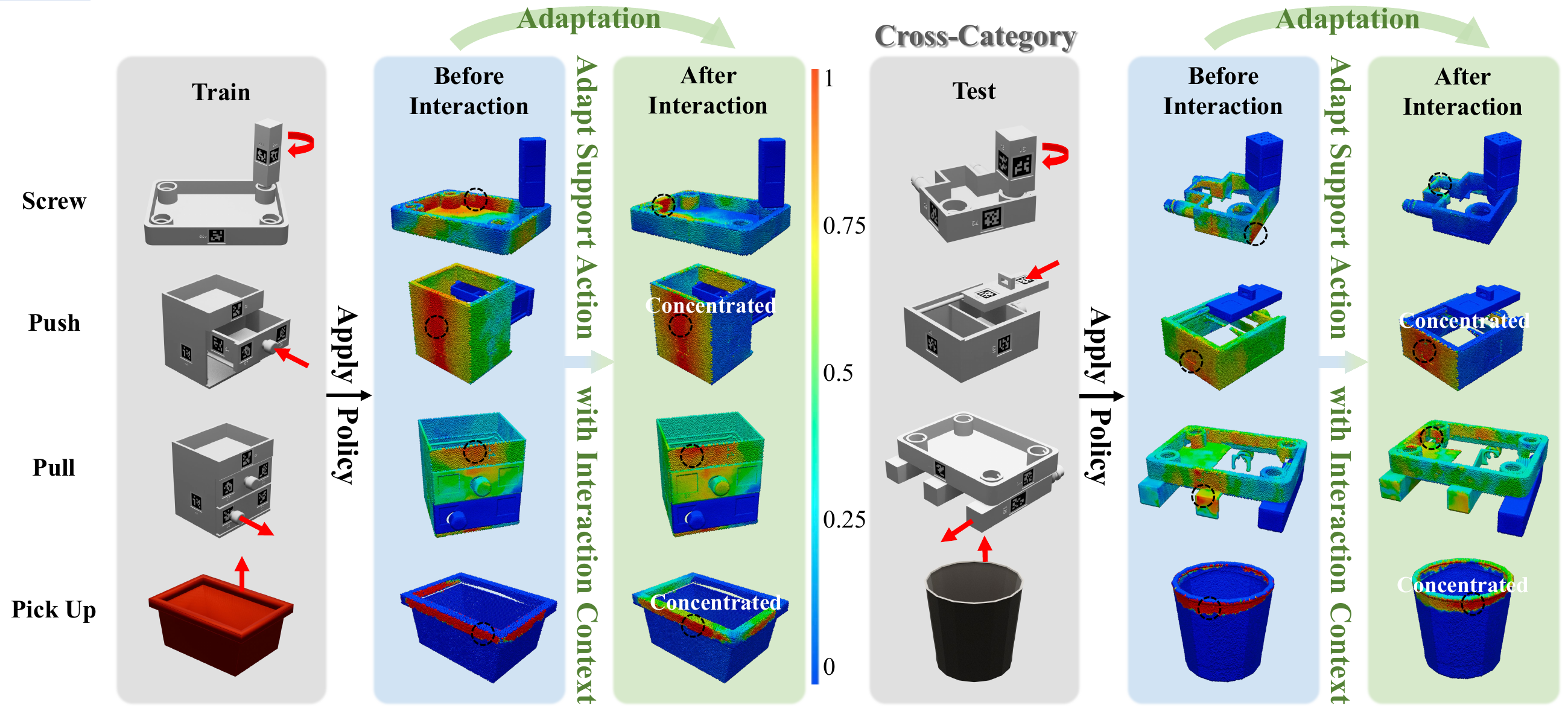}
\caption{\textbf{Affordance Map.} The figure displays affordance heatmaps generated for various objects in simulation before and after interaction. Red arrows indicate the direction of part movement, and circled regions denote the highest-scoring areas. In the subplots labeled “Concentrated,” it is evident that after interaction, high-scoring points converge more tightly at the correct locations; in the other subplots, the high-scoring points have shifted in accordance with the observed movement trends.}
\label{fig:aff}
\end{figure*}

\subsection{Train and Loss}
% In our training process, we first train the Action Scoring Module and Action Proposal Module networks, then the Affordance Prediction Module. The training approach combines both offline data training and online adaptation. Initially, we collect offline data and train the networks using this data to learn the basic task representations. After training the networks with offline data, we proceed to fine-tune them using data generated through interactions between the trained networks and the simulation environment. This online adaptation allows the model to continuously improve based on real-time feedback, optimizing its performance for dynamic task scenarios. The method for collecting training data is detailed in the appendix.

\subsubsection{Action Scoring Loss.  } 
The Action Scoring Module predicts a success score $\hat{r}$ and is trained with an MSE loss against a “real” score $r$ . This real score combines the object’s SE(3) movement distance $g_d$ and a task‐completion term $g_c$—the latter decreasing as completion improves—via weighted sum, then clamps the result to [0,1]:
\begin{equation}
r=\text{clamp}((1-(\alpha\times g_d+\beta\times g_c)), 0, 1)
\end{equation}
where $\alpha$ and $\beta$ balance the distance and completion.

% The MSE loss is then calculated as:
% $$\mathcal{L}_{\text{scoring}}=\frac{1}{N}\sum_{i=1}^{N}(r_i-\hat{r}_i)^2$$
% where $r_i$ is the real score for the i-th interaction, and $\hat{r}_i$ is the predicted success score from the network.

\subsubsection{Action Proposal Loss.  }
% For the training of the Action Proposal Module, we only use data where the operation was successful and the task completion was satisfactory. Since the prediction output is a direction vector, 
We evaluate the loss using cosine similarity and Kullback-Leibler (KL) divergence.
\begin{enumerate}
% \item \textbf{Cosine Similarity Loss}: We use cosine‐similarity loss to align the predicted direction $\hat{\mathbf{d}}$ with the ground-truth $\mathbf{d}$: 
% \begin{equation}
% \mathcal{L}_{\text{cosine}}=1-\frac{\hat{\mathbf{d}} \cdot \mathbf{d}}{\|\hat{\mathbf{d}}\| \|\mathbf{d}\|}
% \end{equation}
% This loss function encourages the predicted direction vector to align as closely as possible with the true direction, minimizing the angle between them.
\item \textbf{Cosine Similarity Loss}: We use a cosine-similarity loss $\mathcal{L}_{\text{cosine}}=1-(\hat{\mathbf{d}} \cdot \mathbf{d}) / (\|\hat{\mathbf{d}}\| \|\mathbf{d}\|)$ to align the predicted direction $\hat{\mathbf{d}}$ with the ground-truth $\mathbf{d}$.
\item \textbf{KL Divergence Loss}: We add a KL‐divergence term to regularize the latent variable $z$ inferred from $\hat{\mathbf{d}}$ and $f^{in}$ towards a standard normal:
\begin{equation}
\mathcal{L}_{\text{KL}}=D_{\text{KL}}(q(z|\hat{\mathbf{d}},f^{in}) || \mathcal{N}(0,1)
\end{equation}
\end{enumerate}
The overall Action Proposal loss then balances direction alignment and latent regularization:
\begin{equation}
\mathcal{L}_{proposal}=\lambda_{dir}\mathcal{L}_{cosine}+\lambda_{KL}\mathcal{L}_{KL}
\end{equation}
where $\lambda_{dir}$ and $\lambda_{KL}$ weight the cosine similarity and KL terms, respectively. 

\subsubsection{Affordance Prediction Loss.  }
Similar to Where2Act and DualAfford, we define each point’s affordance score $a$ as the predicted success probability of actions proposed by the Action Proposal Module, and evaluated by the Action Scoring Module. Concretely, for each point $p_i$ we sample $N$ support directions, score them via Action Scoring Network to obtain $N$ action scores and average the top.
% From these scores, we select the top $K$ highest-rated scores and compute their average to determine the affordance score $a_{p_i}$ for $p_i$. 
\begin{equation}
a_{p_i}=\frac{1}{K}\sum_{j=1}^{K}\mathscr{S}(f^{in}_{pi},\mathscr{P}(f^{in}_{pi},z_j))
\end{equation}
We then apply L1 loss to measure the difference between the predicted affordance score $\hat{a_{p_i}}$ and the ground-truth $a_{p_i}$:
\begin{equation}
\mathcal{L}_{\text{affordance}}=|\hat{a_{p_i}}-a_{p_i}|
\end{equation}

\section{Experiment}
\subsection{Setup}
\paragraph{Environment.} We build upon FurnitureBench in IsaacGym by extending it to support dual-arm coordination and modifying camera configurations, allowing us to study the collaborative support and stabilization using a second arm.
To boost and evaluate policy generalization, we extend the assets by increasing object geometric diversity.
For training, we collect ~10k samples per task focusing on specific furniture types (e.g., desk, drawer, basket), each with multiple variants. Testing utilizes entirely unseen furniture types to validate cross-category generalization.
% such as replacing rectangular tables with triangular and trapezoidal variants. 
% This results in 50+ geometrically diverse parts across 8 furniture types.

\paragraph{Tasks.} We evaluate on four fundamental assembly operations: (1) \textbf{Screwing}: rotating components while the support arm provides counteracting force; (2) \textbf{Insertion}: pushing components along rails with support arm guidance; (3) \textbf{Extraction}: pulling components while the support arm stabilizes the base; (4) \textbf{Picking}: lifting and placing with dual-arm coordination.

\paragraph{Metrics.} We use success rate as the evaluation metric. Success requires: target component reaching desired pose within tolerance, base structure remaining stable (displacement/rotation below thresholds), and secure grasping above specified height for picking tasks.
% We use success rate as the evaluation metric. A trial is considered successful if it satisfies all of the following conditions: the target component reaches the desired pose within a preset tolerance, the main structure remains stable (with displacement and orientation changes below defined thresholds), and, for picking tasks, the object is securely grasped and lifted above a specified height.

\begin{table*}[!ht]
 
  \centering
  \begin{tabular}{lcccc|cccc}
    \toprule
    & \multicolumn{4}{c}{Train Categories} &\multicolumn{4}{c}{Test Categories}\\
    \cmidrule(lr){2-5}\cmidrule(lr){6-9}
    % \multicolumn{2}{c}{Part}                   \\
    % \cmidrule(r){1-2}
    Method      & Screw& Push& Pull & Pick Up & Screw& Push& Pull & Pick Up\\
    \midrule
    Random      & 10.7\%   & 11.1\%  &  5.0\%  & 13.9\% & 9.0\%  & 7.2\%  & 4.0\%  & 6.5\% \\
    Heuristic   & 54.5\%   & 70.9\%  & 43.1\%  & 37.5\%  & 46.7\%  & 52.8\%  & 31.9\%  & 31.4\%\\
    DP3         & 23.2\%   & 41.5\%   & 19.4\%   & 22.9\%  & 17.4\%  & 22.1\%  & 10.1\%  & 11.5\%\\
    LLM-Guided  & 0.0\%    & 0.0\%   & 0.0\%   & 0.0\%  & 0.0\%  & 0.0\%  & 0.0\%  & 0.0\%\\
    w/o Top‑K  
                & 66.7\%  & 73.5\%  & 67.7\% & \textbf{66.1\%} & 54.1\%  & 53.7\%  & 52.4\%  & 41.1\%\\
    w/o Adaptation     
                & 54.9\%  & 74.3\%  & 76.3\%  & 52.9\% & 34.2\%  & 63.9\%  & 55.6\%  & 42.2\%\\
    Ours        & \textbf{70.7\%}  & \textbf{80.6\%}  & \textbf{80.0\%}  & 62.0\%  & \textbf{56.3\%}  & \textbf{67.9\%}  & \textbf{61.7\%}  & \textbf{47.1}\%\\
    \bottomrule
  \end{tabular}
   \caption{Comparison of baseline and ablation variants on the success rate metric.}
  \label{table:baseline}
\end{table*}

\begin{figure*}[!h]
\centering
\includegraphics[width=\textwidth]{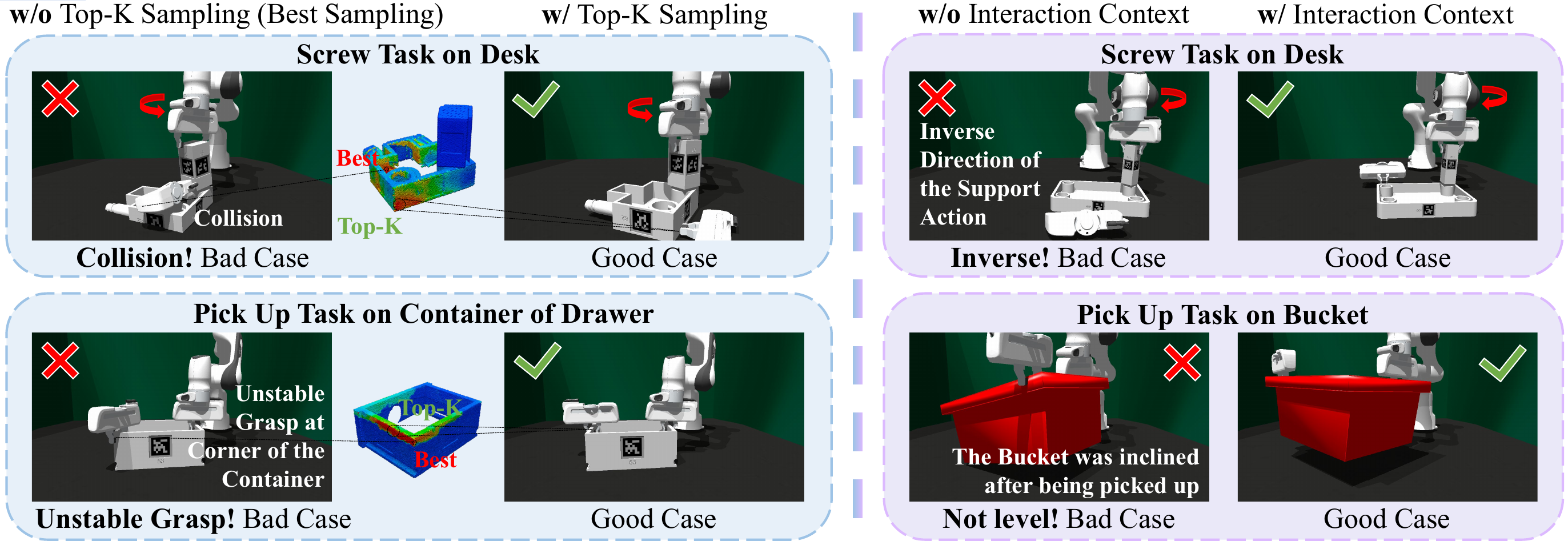}
\caption{\textbf{Qualitative Analysis of Ablations.} (Left) Without Top-K sampling, the robot fails to find robust manipulation points. (Right) Without interaction context, the robot lacks physical awareness to adjust its actions.}
\label{fig:Ablation}
\end{figure*}

\subsection{Baselines and Ablations}
Our work targets adaptive support in dual-arm assembly, a novel setting not directly addressed by prior work. Thus, no existing method serves as a direct SOTA baseline. We compare against the following baselines and ablations:
\begin{itemize}
\item \textbf{Random}: Random selection of support points and directions.
\item \textbf{Heuristic}: Support point selection by geometric rules.
\item \textbf{3D Diffusion Policy (DP3)}~\citep{ze2024dp3}: Imitation learning for support prediction with point cloud input.
\item \textbf{LLM-Guided}~\citep{comanici2025gemini}: Inferring support point and action using Gemini 2.5 Pro.
\end{itemize}
To demonstrate the necessity of the proposed module, we compare with the following ablated versions:
\begin{itemize}
% \item \textbf{w/o Top-K}: Single highest-scoring point instead of multi-point sampling.
\item \textbf{w/o Top-K}: Selecting only the single highest-scoring point instead of Top-K candidates.
\item \textbf{w/o Adaptation}: Removing adaptation with interaction.
\end{itemize}

\subsection{Results and Analysis}

Fig.~\ref{fig:aff} demonstrates the predicted affordance before and after the interactions, for different tasks, over training and novel object categories.
Before the interaction,
the learned affordance might be ambiguous (indicating a larger number of points that are plausible for manipulation) due to the uncertainty of object kinematics and dynamics.
After a support action executed by the second robot, on the point selected by the proposed affordance,
the affordance will be adapted by the interaction feedback.
Eventually,
the manipulation regions indicated by the adapted affordance will be more concentrated on plausible support points.

% Moreover, as shown in Fig.~\ref{fig:aff},
% the learned and adapted affordance, and the corresponding policy can generalize to shapes with different geometries in novel categories,
% as the learning paradigm of point-level affordance aggregates both the low geometry (indicating where can be manipulated) and overall structure (indicating where to support).
Moreover, the learned and adapted affordance, and the corresponding policy can generalize to novel geometries and categories, as point-level affordance aggregates both the low geometry (indicating where can be manipulated) and overall structure (indicating where to support).

% \paragraph{Main Results.} 
Tab.~\ref{table:baseline} shows the quantitative results, and our proposed framework outperforms all baseline and ablation methods.
\textbf{Heuristic} method, though more effective than \textbf{Random} actions, requires manual rule design for each task and even object.
\textbf{DP3} lacks the understanding of diverse shapes and categories. \textbf{LLM-Guided} approaches lack essential 3D geometry and a low-level fine-grained action understanding for precise manipulation.
% The \textbf{Random} strategy, which relies entirely on blind sampling, representing the “lower bound” of the action space. Although \textbf{Heuristic} method can achieve high scores, designing specific heuristic rules for every object shape is not realistic. As a representative of imitation learning, \textbf{DP3} executes demonstration actions solely based on local geometry–action mappings; when deviations occur (e.g., across unseen categories), the mapping fails and the model performs neither exploration nor correction. The \textbf{LLM} only performs semantic-level image–text mapping, lacking 3D geometric understanding from depth or point-cloud data, and therefore cannot succeed in execution. In contrast, \textbf{our method} achieves the highest success rates for all tasks in the training categories and shows only a slight decrease when transferred to unseen categories, strongly demonstrating its effectiveness.

For the analysis of \textbf{ablations}, Tab.~\ref{table:baseline} and Fig.~\ref{fig:Ablation} 
together showcase the effectiveness of the proposed components.
% confirm that both Top-K sampling and the interaction context module are essential. 
% \textbf{Top-K sampling} expands the action search space, preventing failures from suboptimal single-point selections. As Figure~\ref{fig:Ablation} (left) shows, without Top-K sampling, the highest-scoring point may cause object compression or unstable grasps on corners.
% The \textbf{interaction context module} enables physical property inference during execution. Figure~\ref{fig:Ablation} (right) demonstrates failure cases without this module: incorrect screw thread direction perception and inability to adapt to object tilting during pickup. Figure~\ref{fig:aff} shows that interaction context significantly improves affordance prediction by capturing force directions and stability requirements.

\begin{figure*}[!h]
\centering
\includegraphics[width=\textwidth]{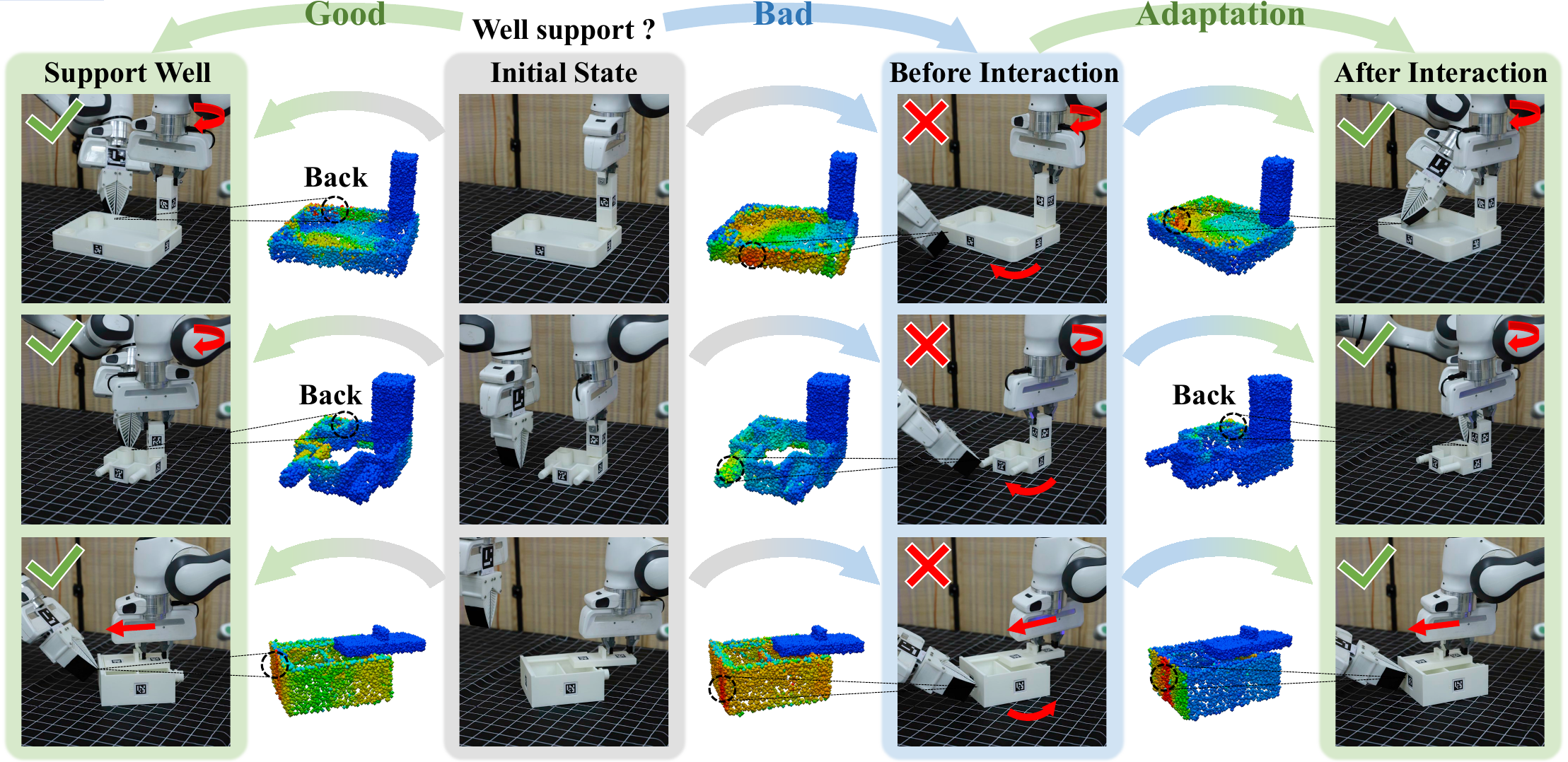}
\caption{\textbf{Real-World Experiments.} We validate our framework in real-world conditions. The experiments include three scenarios:screw a desk leg, screw chair leg, and push a cabinet door. The left path ("Good") shows our policy directly finding a stable support. The right path ("Bad" to "Adaptation") show the ability to adapt its support strategy to ultimately succeed.}
\label{fig:Real-World}
\end{figure*}

For \textbf{Top-K Sampling}, generates a wider set of high-quality candidate actions for the support goal, for the following Action Scoring Module to further select the best actions.
On the contrary,
if the framework only selects the best point indicated by the learned affordance,
chances are that on this selected point the best action direction is worse than the action directions sampled on other points with high affordance scores.
% Therefore,
% a larger set of point candidates with corresponding actions make it easier to select the best support action.
The left side of Fig.~\ref{fig:Ablation} illustrates two failure cases when Top-K sampling is omitted. In the top-left case, although the affordance module provides the highest-scoring contact point, its combination with the direction proposed by the action module fails to provide optimal support due to the collision problem. In the bottom-left case, the highest-scoring contact point is located on a corner of the box that yields an unstable grasp, making successful execution highly improbable. In our scenarios with complex geometries and diverse tasks, \textbf{Top-K sampling} effectively expands the high-quality search space, markedly improving the probability of selecting the best action.

The \textbf{Adaptation Mechanism} based on the interaction context enhances the perception of real-time physical properties, endowing the model with 'physical awareness'. The right side of Fig.~\ref{fig:Ablation} presents failure examples when this module is removed. In the top-right of the "Screw" task, although providing a seemingly correct support action, without interaction contexts, the model is unaware of the complex thread direction. This leads to an inverse action that cannot support the tightening operation well.
In the bottom-right of the "Pick-up" task, the initial grasp causes the bucket to incline; by incorporating this tilt as interaction feedback, the model can automatically adjust the contact point and successfully lift the target object.
As shown in Fig.~\ref{fig:aff}, after incorporating interaction context, the model not only highlights contactable regions more accurately, but also reveals differences in physical interaction properties such as force direction and stability, significantly enhancing its perception and understanding of interaction states.

\subsection{Real-Word Experiments}

\begin{table}[htbp]
  \centering
  \begin{tabular}{lcccc}
    \toprule
    Method    & Screw & Push  & Pick Up \\
    \midrule
    Random    & 0 / 15  & 0 / 15 & 0 / 15    \\
    Heuristic & 8 / 15  & 10 / 15 & 5 / 15    \\
    DP3       & 4 / 15  & 6 / 15 & 5 / 15    \\
    Ours      & \textbf{11 / 15} & \textbf{12 / 15} & \textbf{9 / 15} \\
    \bottomrule 
  \end{tabular}
  
  \caption{Real world experimental results.}
  \label{table:real_word}
\end{table}

We set up two Franka Panda with the furniture positioned between them. Three RealSense cameras capturing 3D point cloud are mounted around the scene. Robot control is managed through ROS~\citep{quigley2009ros} and the frankapy library~\citep{zhang2020modular}. Fig.~\ref{fig:Real-World} demonstrates the complete pipeline from scene perception and adaptive affordance prediction based on interaction feedback.

We evaluate each task over 15 trials with varying furniture configurations on 3 tasks. As shown in Tab.~\ref{table:real_word}, our method significantly outperforms baselines and achieves high success rates in real-world assembly tasks. 
Fig. ~\ref{fig:Real-World} shows real-world observations, affordance and adaptation. Additional videos are provided in the supplementary material.

The primary failure mode in real-world stems from motion planning limitations.The RRTConnect algorithm cannot find feasible trajectories due to robotic arm or environmental constraints. In the future work, we plan to develop a policy for motion refinement to improve real-world robustness.

\section{Conclusion}
We propose A3D, a framework that learns adaptive affordances for dual-arm furniture assembly by identifying optimal support and stabilization locations. Our approach combines dense geometric representations for cross-geometry generalization with an adaptive module that leverages interaction feedback to dynamically adjust strategies. Experiments demonstrate superior performance in both simulation and real-world settings.

\section{Acknowledgments}
The authors gratefully acknowledge the hard work and close collaboration of the team members. In particular, Jiaqi contributed to the code implementation, experimental work, and real-robot experiments; Yue provided mentoring and contributed to experimental design, real-robot experiments, and manuscript preparation; Qize contributed to portions of the experiments, including real-robot experiments, as well as figure preparation and website development. Yan assisted with manuscript polishing, and Ruihai contributed to idea generation and overall research advising.Haipeng initialized the project code and simulation. 

The funding author for this paper unexpectedly withdrew the funding after camera-ready submission, and Ruihai covered all the associated costs.
\bibliography{aaai2026}

% --- 开始附录，新启一页 ---
\clearpage
\appendix

% --- Appendix A ---
\section{Simulation Environment}
\label{app:simenv}

\subsection{FurnitureBench}
FurnitureBench is a benchmark designed for real-world furniture assembly tasks. Built on Isaac Gym, it provides a single-arm Franka robot and a variety of furniture assets. We extended the original simulation environment with several modifications: we introduced an auxiliary gripper, designating the existing Franka arm as the primary manipulator and the added gripper as an assistant to execute our support policy.

Concurrently, we adjusted the number and placement of cameras in the environment by deploying four cameras—positioned at the front, rear, left, and right—to capture depth images, which were subsequently processed into point clouds. The camera resolution is set to $1280 \times 720$ with a frame rate of 30 fps.

\begin{figure}[!ht]
\centering
\includegraphics[width=0.4\textwidth]{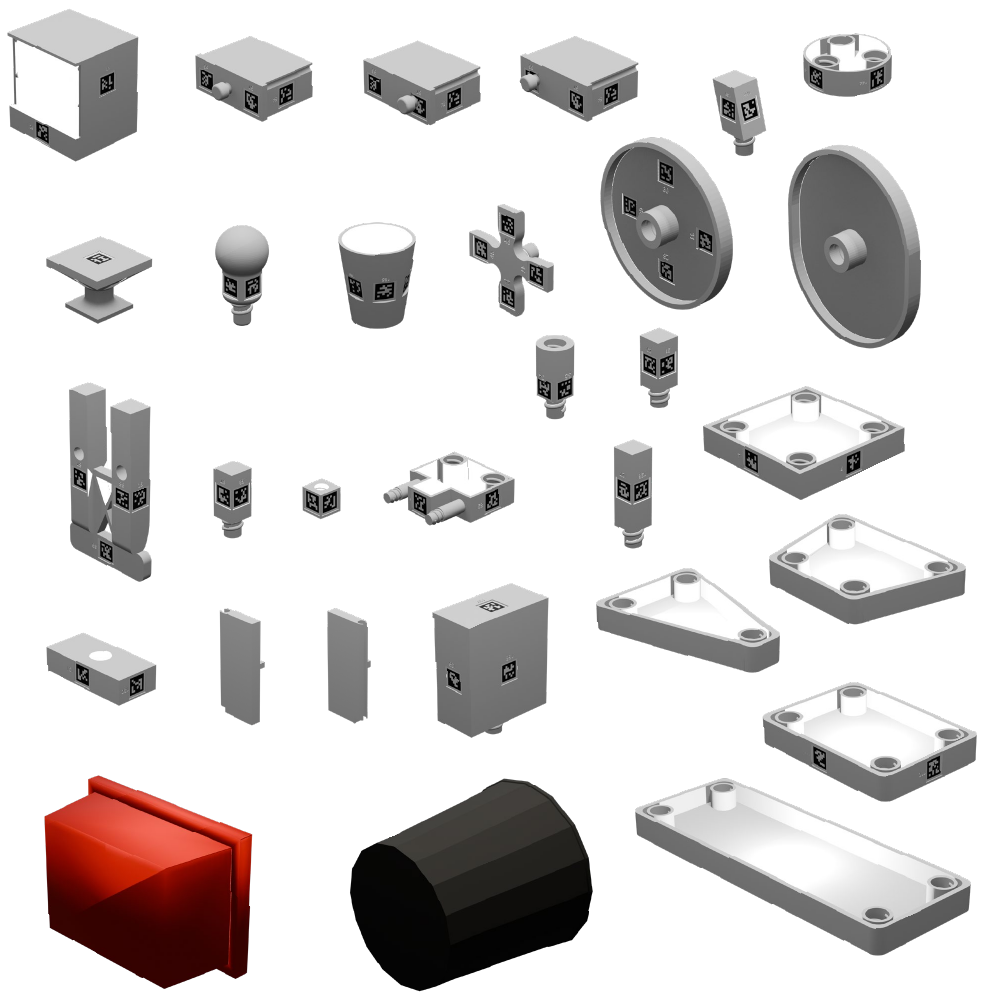}
\caption{Partial Assets utilized in the simulation.}
\label{fig:p_assets}
\end{figure}

Additionally, to validate our method's generalization across different geometries, we manually adjusted the meshes of selected furniture assets, creating a broader set of parts with diverse shapes.

\subsection{Task Details}
We selected four common operations in assembly scenarios as our primary task categories:
\begin{itemize}
    \item \textbf{Screwing}: This category includes specific tasks such as screwing legs onto tables of various shapes, tightening chair legs, and screwing bulbs into lamps.
    \item \textbf{Pushing}: This category includes tasks such as inserting drawer boxes into drawer bodies at various handle positions and installing cabinet doors into wardrobe frames along guide rails.
    \item \textbf{Pulling}: This category includes pulling out a drawer box and extracting target components from a stacked assembly.
    \item \textbf{Picking Up}: This category involves picking up objects of various shapes, such as drawer boxes, buckets, baskets, and boxes.
\end{itemize}

All tasks inherently require dual-arm coordination: the primary arm executes the operation (screwing, pushing, pulling, or picking), while the support arm applies stabilizing forces to ensure physical feasibility. Without this support policy, these tasks cannot be completed successfully in either simulation or real-world settings.

% --- Appendix B ---
\section{Real World Experiments}
\label{app:realworld}

% TODO: Insert figure of real-world experiment environment (showing all assets) here if available.

In the real-world experiments, we evaluated the \textit{screw}, \textit{push}, and \textit{pick up} tasks. Due to the size constraints of the 3D-printed parts and the physical workspace limitations of the Franka arm, we did not perform experiments for the \textit{pull} task. The task configurations remained consistent with those in the simulation.

Notably, since the LLM-guided approach performed poorly in the simulation environment, it was not included in the real-world experimental comparisons.

% --- Appendix C ---
\section{Simulation Experiments}
\label{app:sim_exp}

For each method, we conducted three test rounds consisting of 100 trials each and reported the average success rate.

\subsection{LLM-guided Baseline}
\subsubsection{Implementation Details}
The LLM baseline receives four RGB views and task-specific textual prompts. Although the model allows for coarse-level object localization, its predictions frequently fall off-object or violate physical constraints (e.g., selecting non-contactable surfaces). Consequently, the success rate was 0\% across all tasks. We retain this baseline primarily for reference to highlight the gap between geometric/physical reasoning and current vision–language capabilities.

\subsubsection{Prompt Design}
The box below illustrates the prompt structure used. For different tasks, \texttt{\{object\_name\}} and \texttt{\{task\_specific\_description\}} are replaced with their respective task-specific content.

\begin{framed}
You will receive four RGB images showing a robotic workspace from the front, back, left, and right perspectives, in that order. The scene depicts a human hand (simulated by another robot) and a robot arm working together to assemble a \{object\_name\}. The current task is to \{task\_specific\_description\}.

First, you must identify the precise location and boundaries of the target object in each image—remember it typically has a color that contrasts strongly with the background and is usually near the center of the frame. Once you have localized the object, choose the single best interaction point on its surface.

Your goal is to identify the single best point for the robot arm to interact with to successfully assist the human hand. Analyze all four images for full context, but your final output must be a three-element tuple indicating:
1. Which image (use 0 for front, 1 for back, 2 for left, 3 for right)
2. The row index (pixel row)
3. The column index (pixel column)

The chosen point must lie on the target object and represent a stable, effective location for the robot's gripper to apply force or grasp. Double-check that your selected point truly fulfills the task's requirements before answering.

You must return **only** the 3-tuple. No additional text or explanation.

For example: (0, 412, 351)
\end{framed}

Below is an example of a task-specific description used in the prompt:
\begin{framed}
\textbf{Screw Task Description:}
Assist with a screwing task. The human hand is actively screwing a part (e.g., a leg) into the main body of the \{object\_name\}. To prevent the main body from rotating, the robot arm needs to apply counter-pressure by firmly holding or bracing the main body. Identify the best point on the main body for the robot to press against to provide this stability.
\end{framed}

\subsubsection{Results}
\begin{figure}[!ht]
\centering
\includegraphics[width=0.4\textwidth]{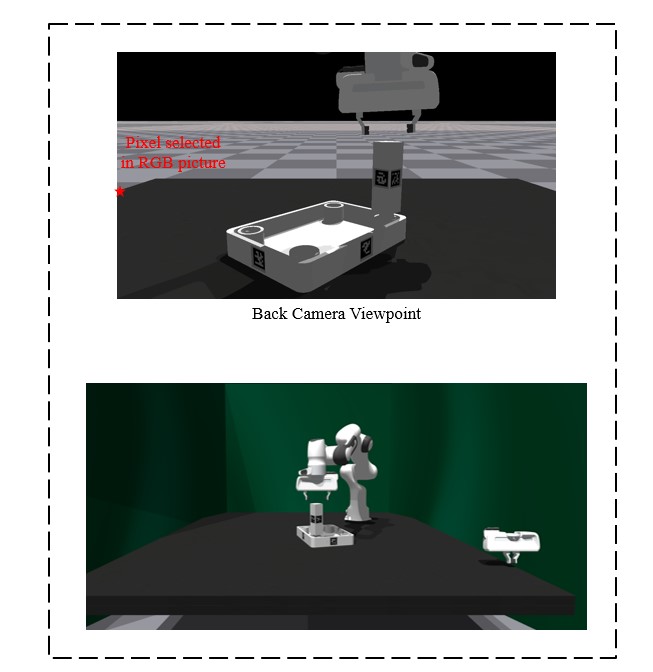}
\caption{An example of an LLM failure case.}
\label{fig:llm_failed}
\end{figure}

Fig.~\ref{fig:llm_failed} illustrates a failure case where the selected point does not lie on the object surface.

% --- Appendix D ---
\section{Implementation Details}
\label{app:imp_details}

\subsection{Data Collection}

\subsubsection{Dataset Overview}
For each task category (screwing, pushing, pulling, picking), we collected approximately 10,000 samples on a single furniture type (desk for screwing, drawer for push/pull, basket for pick), each containing several geometric variants. All evaluation tasks utilized unseen furniture types to ensure category-level generalization.

\subsubsection{Offline Data Collection}
During the data collection process, we observed that traditional random sampling was inefficient; some tasks yielded very low success rates with random sampling, hindering the effective training of the Action Proposal Module. To address this, we adopted a hybrid sampling approach combining random sampling and heuristic sampling. 

In \textit{random sampling}, we select a point randomly and sample a support direction near the point's normal. In \textit{heuristic sampling}, we manually designate regions on the object's point cloud where interaction is more likely to succeed for specific tasks. We then sample points and support directions within these regions. This hybrid approach ensures both diversity and a sufficient proportion of successful trials for effective training. Crucially, the collected offline data includes both pre-interaction successes and failures, enabling the model to learn a robust prior affordance distribution before online adaptation.

Since our model requires \textbf{interaction context} data, we employed a multi-step interaction sampling strategy. We define a maximum interaction context horizon and a movement threshold $\epsilon$. If the manipulated object's movement exceeds $\epsilon$, we infer that the current support action $u$ is insufficient. We then record the action $u$, the pre-operation point cloud $O$, and the movement $m$ as one interaction context frame. During multi-step sampling, we randomly switch between random and heuristic methods at each step until task completion or the horizon limit is reached. We collected 10,000 samples for each task to train our policy.

\subsubsection{Online Adaptive Data Collection}
Although the hybrid offline collection improves efficiency, it may not cover all scenarios. To enhance robustness, we propose online adaptive data collection to sample from a broader range of potential success sub-regions.
Specifically, the model selects the most suitable manipulation point and generates action proposals. The best proposal is selected via a scoring mechanism, and action $u$ is executed. During execution, multiple actions may be chosen (similar to offline collection) to generate interaction context data. Upon task completion, the data is saved.

When the collected online data reaches a specified volume, it is combined with an equal number of offline samples. This mixed dataset is used to fine-tune the model.
For adaptation, we use 64 samples (32 from the current test interaction, 32 from prior offline data) per update, repeating the update 3 times. This is an offline adaptation process rather than continuous online fine-tuning. Inference speed remains at the millisecond level, causing no noticeable delay.

\subsection{Policy Network}
The network architecture is detailed in the main text. Specifically, we employ a segmentation version of PointNet++ as the visual feature extractor, obtaining 128-dimensional features for each point. For both action-point and support-point features, we use the same single-layer MLP to extract 32-dimensional representations; support-direction features are similarly extracted via a single-layer MLP with 32-dimensional outputs. The Affordance and Critic modules are implemented as multi-layer MLPs (hidden dim=128) mapping input features to scores. The Actor module utilizes a Conditional Variational Autoencoder (CVAE) with hidden layers and latent space dimension set to 128. Each module’s feature extractor is independent and does not share parameters.

\subsection{Training Details}
For the Critic module, we set the weights $\alpha$ and $\beta$ of the Action Score Loss to 1:1. For the Actor module, the weights balancing KL divergence ($\lambda_{KL}$) and cosine similarity ($\lambda_{dir}$) in the Action Proposal Loss are also set to 1:1.

We employ the \textbf{Adam} optimizer with an \textbf{initial learning rate} of \textbf{0.001} and a \textbf{weight decay} of \textbf{1e-5}. A \textbf{StepLR scheduler} multiplies the learning rate by \textbf{0.9} every \textbf{500} steps. The \textbf{batch size} is \textbf{64}. We use early stopping if the learning rate falls below 5e-7.

\subsection{Software and Hardware}
Experiments were conducted on Ubuntu 20.04.6 LTS, equipped with an Intel(R) Xeon(R) Gold 5220 CPU @ 2.20 GHz and 125 GiB RAM. The GPU is an NVIDIA GeForce RTX 3090 with CUDA 12.4.

\end{document}